\documentclass[11pt,letterpaper]{article}
\usepackage{ijcnlp2017}
\usepackage{times}
\usepackage{latexsym}
\usepackage{array}
\usepackage{caption}
\usepackage{multirow}
\usepackage{graphicx}
\usepackage{url}

\usepackage[font=small,labelfont=bf]{caption}
\DeclareCaptionFont{tiny}{\tiny}

\ijcnlpfinalcopy

\def\ijcnlppaperid{16}

\title{``\textit{Having 2 hours to write a paper is fun!}'': Detecting Sarcasm in Numerical Portions of Text}

 \author{Lakshya Kumar, Arpan Somani, Pushpak Bhattacharyya \\
  Dept. of Computer Science and Engineering\\
  IIT Bombay, India\\
  {\tt lakshya,somani,pb@cse.iitb.ac.in}}

\date{}

\begin{document}

\maketitle

\begin{abstract}
 Sarcasm occurring due to the presence of numerical portions in text has been quoted as an error made by automatic sarcasm detection approaches in the past.
We present a first study in detecting sarcasm in numbers, as in the case of the sentence `\textit{Love waking up at 4 am}'. 
We analyze the challenges of the problem, and present Rule-based, Machine Learning and Deep Learning approaches to detect sarcasm in numerical portions of text. Our Deep Learning approach outperforms four past works for sarcasm detection and Rule-based and Machine learning approaches on a dataset of tweets, obtaining an F1-score of $0.93$. This shows that special attention to text containing numbers may be useful to improve state-of-the-art in sarcasm detection.
\end{abstract}

\section{Introduction}
Computational detection of sarcasm has seen attention from the sentiment analysis community in the past few years \cite{joshi2016automatic}. Sarcasm is an interesting problem for sentiment analysis because surface sentiment of words in a sarcastic text may be different from the implied sentiment. For example, `\textit{Being stranded in traffic is the best way to start a week}' is a sarcastic sentence because the surface sentiment of the word `\textit{best}' (positive) is different from the implied sentiment of the sentence (negative), considering remaining portions of the text.

Past approaches for sarcasm detection report features related to sentiment \cite{gonzalez2011identifying}, author's historical context \cite{rajadesingan2015sarcasm}, and conversational context \cite{joshi2016harnessing}. Error analysis presented in many of these works has served as a motivation for future work. Our paper is based on an error observed by~\citet{joshi2015harnessing}: `Incongruity in numbers, resulting in sarcasm'. Consider the sentence in the title of this submission: `\textit{Having 2 hours to write a paper is fun\footnote{This sentence is only an example. This paper was not written in 2 hours.}}'. The number `\textit{2 hours}' is a crucial indicator of the sarcasm in this sentence. The sarcasm is based on the understanding that two hours may not be sufficient to write a  paper. If the number of hours is changed to a higher value, the sarcasm in the sentence may not hold, given that the value is a sufficient duration to write a paper. This paper deals with detecting sarcasm in numerical portions of text, as in this example.

We first  present a rule-based approach to detect sarcasm expressed due to numbers. Our approach compares numerical magnitudes with those seen in similar contexts in a training dataset. Since `\textit{similar context}' is key here, we consider two variants of our approach in order to match the context. Then we present Machine learning based approach and its variant that take different features as input for learning. Further we propose deep learning based approaches to numerical sarcasm detection on social media that does not require extensive manual feature engineering. Instead, we develop Convolution Neural Network (CNN) to capture local correlations of spatial or temporal structure. We also develop Long-short Term Memory (LSTM) network which is able to handle sequences of any length and capture long-term dependencies. We compare our approaches with four past works, and show an improvement. To the best of our knowledge, this is the first reported work that deals with sarcasm in numerical portions of text. We present our evaluation on a dataset of tweets.

The rest of the paper is organized as follows. We describe the  related work in Section~\ref{sec:relwork}. Then, we motivate this work in Section~\ref{sec:motiv}, and describe various approaches to detect sarcasm in numerical portions of text in Section~\ref{sec:approaches}. The experiment setup is outlined in Section~\ref{section:expsetup}. Results of our experiments are given in Section~\ref{sec:res}, while Section~\ref{sec:erroranal} discusses the errors.  
Finally, the conclusion and future work is described in Section~\ref{sec:concl}.

\section{Related Work}
\label{sec:relwork}
Sarcasm and irony detection has been extensively studied in linguistic, psychology and cognitive science \cite{gibbs1986psycholinguistics,utsumi2000verbal}.
Computational detection of sarcasm has become a popular area of natural language processing research in recent years \citet{joshi2016automatic}.
\citet{tepperman2006yeah} present sarcasm recognition in speech using spectral (average pitch, pitch slope, etc.), prosodic and contextual cues.
\citet{carvalho2009clues} use simple linguistic features like interjection, changed names, etc. for irony detection.
\citet{davidov2010semi} train a sarcasm classifier with syntactic and pattern-based features.
\citet{gonzalez2011identifying} states that sarcasm transforms the polarity of an apparently positive or negative utterance into its opposite. \citet{liebrecht2013perfect} showed that sarcasm is often signaled by hyperbole,  using intensifiers and exclamations; in contrast, non-hyperbolic sarcastic messages often receive an explicit marker.

\citet{riloff2013sarcasm} states that sarcasm is a contrast between a positive sentiment word and a negative situation. \citet{buschmeier2014impact} provided the baseline for classification of ironic or sarcastic reviews. They analyzed the impact of different features for the classification task. The work by \citet{joshi2015harnessing} shows how sarcasm arises because of implicit or explicit incongruity in the sentence. \citet{bouazizi2016pattern} proposed a pattern-based approach to detect sarcasm on Twitter. They proposed four sets of features that cover the different types of sarcasm. 

As deep learning techniques gain popularity, few deep learning based architectures for sarcasm detection have also appeared in literature. \citet{ghosh2016fracking} provides a neural network semantic model for sarcasm detection. Their model composed of Convolution Neural Network (CNN) followed by a Long Short Term Memory (LSTM) network and finally a Deep Neural Network(DNN).
\citet{poria2016deeper} proposed a novel method to detect sarcasm using Convolution Neural Networks. They have developed models based on a pre-trained convolutional neural network for extracting sentiment, emotion and personality features for sarcasm detection. \citet{amir2016modelling} proposed a deep-learning based architecture to automatically learn user embeddings. In their proposed approach they have used this user embeddings to provide \textit{contextual features}, going beyond the lexical and syntactic cues for sarcasm.
\citet{zhang2016tweet} used a bi-directional gated recurrent neural network followed by a pooling neural network to detect sarcasm. 

All these past works deals with the detection of sarcasm that occurs in text but they did not talk about incongruity due to numbers. Our work is the first in this area that tackles the task to identify the presence of numerical sarcasm in tweets.

\section{Motivation}
\label{sec:motiv}
Consider the following sentences:
\begin{enumerate}
\setlength \itemsep{0em}
\item \textit{This phone has an awesome battery back-up of 38 hours. (Non-sarcastic)}.
\item \textit{This phone has an awesome battery back-up of 2 hours. (Sarcastic)}.
\item \textit{This phone has a terrible battery back-up of 2 hours (Non-sarcastic)}.
\end{enumerate}

Sentences 1 and 3 are non-sarcastic while Sentence 2 is sarcastic. Consider Sentences 1 and 2. The two sentences differ only in the numerical value (`\textit{38}' versus `\textit{2}'). The sarcasm in Sentence 2 can be understood in terms of the incongruity\footnote{\citet{ivanko2003context} describe the relationship between incongruity and sarcasm.} between the word `\textit{awesome}' and `\textit{2 hours}' in case of the battery life. On the contrary, Sentences 2 and 3 differ in one word with varying surface sentiment (`\textit{awesome}' versus `\textit{terrible}'). Detecting sarcasm in sentences like Sentence 2 using information from Sentences 1 and 3 in a dataset is the key idea of our rule-based approach to detect sarcasm in numerical portions of text.

Since sarcasm is an infrequent phenomenon and we deal with a specific form of sarcasm (namely sarcasm in numerical text), it is worthwhile to estimate how many sarcastic sentences contain numbers. A set of approximately 100,000 sarcastic tweets contained 11,488 tweets with numbers in them, amounting to 11.48\%.


\section{Approaches}
\label{sec:approaches}
In order to detect numerical sarcasm, we implement two rule-based approaches as described in the forthcoming subsections. 
\subsection{Approach-1: Noun phrase exact matching}
In this approach, we first create two repositories, i.e., sarcastic and non-sarcastic using a training dataset. Each entry in the repository is of the format: \\
(\textbf{Tweet Index No., Noun Phrase list, Mean of Number unit\footnote{The average value of all the numbers corresponding to the Number unit}, Std Dev of Number unit\footnote{Standard Deviation of all the numbers corresponding to the Number unit}, Number Unit\footnote{We calculate the Mean and SD for all possible units present in the dataset})}.\\ 
The repositories are created as follows. For each sarcasm-labeled tweet in the training dataset, we perform the following steps:
\begin{itemize}
\setlength\itemsep{0em}
\item  \textbf{Step-1:} Extract noun phrases in the tweet, using a parser.
\item  \textbf{Step-2:} Select Number unit as the word following the word POS tagged as 'CD'. Examples of number units are minutes, hour, days, years etc.
\item  \textbf{Step-3:} We add an entry to the corresponding repository according to the label of the tweet. This entry is of the format as specified above.
\end{itemize}
For example, if there is a sarcastic tweet- ``This phone has an awesome battery back-up of 2 hours", constituency parse tree of this tweet obtained from nltk parser is shown in Figure \ref{fig:GS} :
\begin{figure}[!htb]
\centering
\includegraphics[width=5cm, height=5cm]{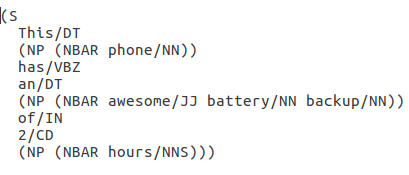}
\caption{Constituency Parse Tree}
\label{fig:GS}
\end{figure}

We extract the Noun phrases from this constituency parse tree and create a noun-phrase list of this sarcastic tweet as follows:\\ 
\textbf{[`phone', `awesome', `battery', `backup', `hours']}\\
After obtaining the noun phrase list, tweet is stored in the sarcastic repository as:\\
\textbf{(Tweet Index No., [`phone', `awesome', `battery', `backup', `hours'], mean of numbers having unit as `hours', Std dev of numbers having unit as `hours' ,`hours' )}\\
Similarly, for all the other tweets, i.e, numerical-sarcastic/ non-sarcastic tweets, same approach is used to store them in their respective repositories. 

Numerical Sarcasm in a test tweet is predicted as follows. We extract noun phrases, number and number unit from the test tweet. Then, following rules are applied:
\begin{itemize} 
\setlength\itemsep{0em}
\item We first consult the sarcastic tweet repository. On matching words in the noun phrase list between the test tweet and entries in the repository, the most similar entry is selected from the sarcastic repository. We then match the number unit of the entry with that in the test tweet.
\item If the number unit matches, we check whether the number present in the test tweet lies within $\pm2.58$\footnote{z value of 2.58 indicates the $99$\% Confidence Interval} Std dev of the mean value for that number unit present in the matched sarcastic entry. For example: \textbf{Test Tweet:} `I love writing this paper at 11 am'. \textbf{Matched Sarcastic Tweet:} `I love writing this paper daily at 3.5 am'. So, the number 11 is not within desired confidence interval and the test tweet is non-sarcastic. 
\item If the number unit does not match,  we consult the non-sarcastic tweet repository and find the most similar entry based on exact matching of the noun phrase list of test tweet and entry in the repository. If the number unit matches, we check whether the number present in the test tweet lies within $\pm2.58$ Std dev of the mean value for that number unit present in the matched non-sarcastic entry. If it lies then predict the tweet as non-sarcastic and if it does not then predict sarcastic. For example: \textbf{Test Tweet:} `I am so productive when my room is 81 degrees'. \textbf{Matched Non-sarcastic Tweet:} `I am very much productive in my room as it has 21.27 degrees'. So, the number 81 is not within desired confidence interval and the tweet becomes sarcastic. 
\item Finally, if no match is found, then the tweet is predicted as non-sarcastic.
\end{itemize}
\subsection{Approach-2: Noun phrase cosine similarity matching }
Approach-1 is restrictive in terms of the phrase matching. Therefore, Approach-2 relaxes the constraint by using cosine similarity to match the phrases in the test tweet and the repository entry. This approach also creates two repositories, i.e., sarcastic and non-sarcastic as follows: 
\begin{itemize}
\setlength\itemsep{0em}
\item  \textbf{Step-1:} We first convert the noun phrase list into its vector representation.
\item  \textbf{Step-2:} This vector representation is created by summation of the 200-dimension word embeddings\footnote{Learned from a tweet corpus containing 6 million tweet words.} of the words present in the noun phrase list and then dividing it by their count.
\item  \textbf{Step-3:} Now the repositories are of the form:\\  (\textbf{Tweet Index No., Vector representation of Noun phrase list,  Mean of Number unit, Std Dev of Number unit, Number Unit}).
\end{itemize}
For example, if there is a numerical sarcastic tweet- ``8:30 am meetings are the best way to start birthday weekend". The Noun phrase list of this tweet is : \\ \textbf{[`meetings', `way', `birthday', `weekend']}\\
In order to create the vector representation of this tweet, 200-dimension vector embedding of each of the word in the noun-phrase list is added and then each component of the resulting embedding is divided by 4 as their are four words in the noun-phrase list. The intuition behind using the vector representation created from the noun-phrase list is that it helps to easily capture the tweets that are similar to the new tweet.  
The rules that are used in this approach are also same as used in Approach-1 but now to match the test tweet and tweets from repository, cosine similarity is used. In this case, the entry with the maximum cosine similarity is selected (as against the entry with exact match in Approach-1).

\subsection{Machine Learning based approach}
\label{ML}
In order to create machine learning based approach for detecting numerical sarcasm we train SVM, KNN and Random Forest classifiers  using different types of features as described below : 
\begin{itemize}
\item \textbf{Sentiment-based features} : These features include number of positive words, number of negative words, number of highly emotional positive words, number of highly emotional negative words. Positive/Negative word is said to be highly emotional if its part-of-speech tag is one among the following :
 `JJ', `JJR', `JJS', `RB', `RBR', `RBS', `VB', `VBD', `VBG', `VBN', `VBP', `VBZ'. 
 \item \textbf{Emoticon-based features} : These features includes positive emoticon, negative emoticon, contrast between word, i.e, a boolean feature that will be one if both positive and negative words are present in the tweet, contrast between emoji, i.e, it will take the value as one when either positive word and negative emoji is present or negative word and positive emoji is present in the tweet. 

\item \textbf{Punctuation-based features} : These features include number of exclamation marks, number of dots, number of question mark, number of capital letter words, number of single quotations.

\item \textbf{Number in the tweet} : This feature is simply the number present in the tweet.

\item \textbf{Number unit in the tweet} : This feature is a one hot representation of the type of unit present in the tweet. Example of number unit can be hour, minute, etc. So, based on the unit present in the tweet that position in the one hot vector takes the value as one and the rest takes the value zero. 

\item \textbf{Tweet Embeddings}: We learn word embeddings of different dimensions, i.e., $25$-D, $50$-D, $100$-D, $150$-D, $200$-D, $250$-D, $300$-D of tweet words using word2vec \cite{mikolov:2013} tool on a large corpora of $6$ million tweets. These word embeddings are used to create the tweet embedding by summing up the word vectors of all the words present in the tweet and finally dividing each component of the resulting vector by the count of total number of words present in the tweet. Finally, we obtain the tweet embeddings of different dimension that we use to train the classifiers. 
\end{itemize}
\subsection{Deep Learning based approach}
\label{DL}
\subsubsection{CNN-FF Model}
\label{cnn_ff}
\begin{figure}[!htb]
\centering
\includegraphics[width=7cm, height=5.5cm]{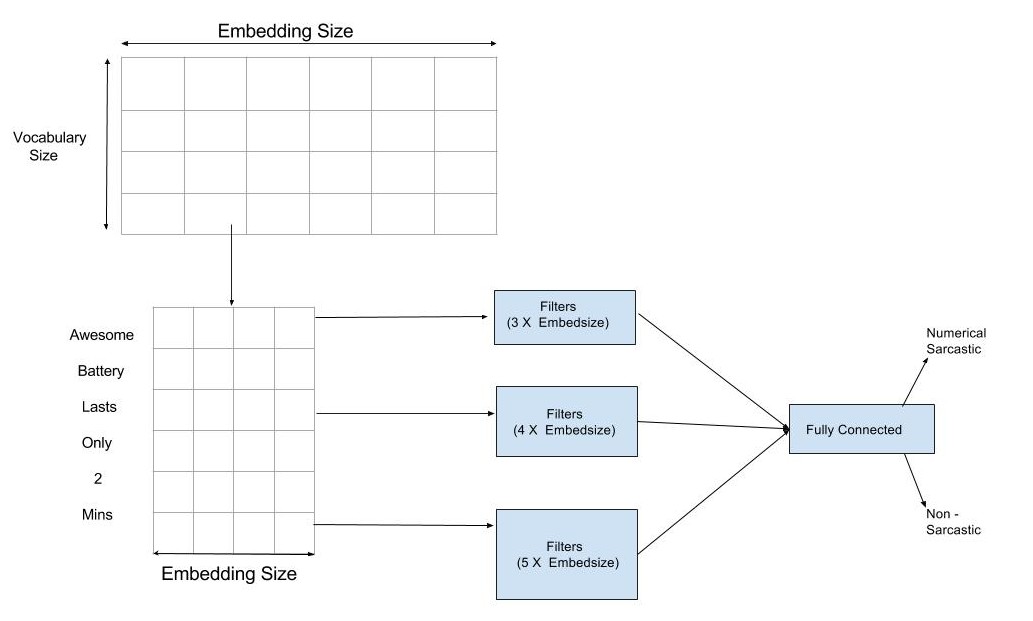}
\caption{CNN followed by Fully Connected}
\label{fig:cnn_ff}
\end{figure}

The architecture of the CNN-FF model is shown in \autoref{fig:cnn_ff}. There is an embedding matrix $E \in {\rm I\!R}^{|V| \times d}$ where $|V|$ is the vocabulary size and $d$ is the tweet word embedding dimension. For the input tweet we obtain an input matrix  $I \in {\rm I\!R}^{|S| \times d}$ where $|S|$ is the length of the tweet including padding, where $I_i$ be the d-dimension vector for i-th word in the tweet in the input matrix.  Let $k$ be the length of the filter, and the vector $f \in {\rm I\!R}^{|k| \times d}$ is a filter for the convolution operation. For each position $p$ in the input matrix I, there is a window $w_{p}$ of $k$ consecutive words, denoted as:
\begin{equation}
w_p = [I_p,I_{p+1},...,I_{p+k-1}]
\end{equation}
A filter $f$ convolves with the window vectors $(k$-grams) at each position in a valid way to generate a feature map $c \in {\rm I\!R}^{|S|-k+1}$ each element $c_p$ of
the feature map for window vector $w_p$ is produced
as follows:
\begin{equation}
c_p = func(w_p \circ  f+b)
\end{equation}
where $\circ$ is element-wise multiplication, $b \in {\rm I\!R}$ is a bias term and $func$ is a nonlinear transformation function that can be sigmoid, hyperbolic tangent, etc. Max-over-time pooling is then applied over the obtained feature map. We use multiple filters of different sizes and output from each filter is concatenated to get the final overall feature vector. This feature vector act as input for the fully-connected layer. We train the entire model by minimizing the binary cross-entropy loss over a mini-batch of training examples of size e.  
\begin{equation}
E(y,\widehat{y}) = \sum\limits_{i=1}^{e} y_i\log(\widehat{y_i})
\end{equation}
\subsubsection{LSTM-FF Model}
RNN have demonstrated the power to capture sequential information in a chain-like neural network. Standard RNNs becomes unable to learn  long-term dependencies as the  gap  between  two  time  steps  becomes  large. We adopted the standard architecture of LSTM proposed by \cite{lstm_hochreiter}.

The LSTM architecture has a range of repeated modules for each time step as in a standard RNN. At each time step, the output of the module is controlled by a set of gates in ${\rm I\!R}^{d}$ as a function of the old hidden state $h_{t-1}$ and the input at the current time step $x_t$: the forget gate $f_t$, the input gate $i_t$, and the output gate $o_t$. These gates collectively decide how to update the current memory cell $c_t$ and the current hidden state $h_t$ . We use $d$ to denote the memory dimension in the LSTM and all vectors in this architecture share the same dimension.  The LSTM transition functions are defined as follows:
\begin{equation}
 i_t = \sigma(W_i \cdot [h_{t-1},x_t]+b_i)
 \end{equation}
 \begin{equation}
  f_t = \sigma(W_f \cdot [h_{t-1},x_t]+b_f)
 \end{equation}
  \begin{equation}
  \widetilde{C_t} = \tanh(W_C \cdot [h_{t-1},x_t]+b_C)  
 \end{equation}
  \begin{equation}
  C_t = f_t \odot C_{t-1} + i_t \odot \widetilde{C_t}
 \end{equation}
  \begin{equation}
  o_t = \sigma(W_o \cdot [h_{t-1},x_t]+b_o)
 \end{equation}
 \begin{equation}
 h_t = o_t \odot \tanh(C_t)
 \end{equation}

\begin{figure}[!htb]
\centering
\includegraphics[width=7cm, height=6cm]{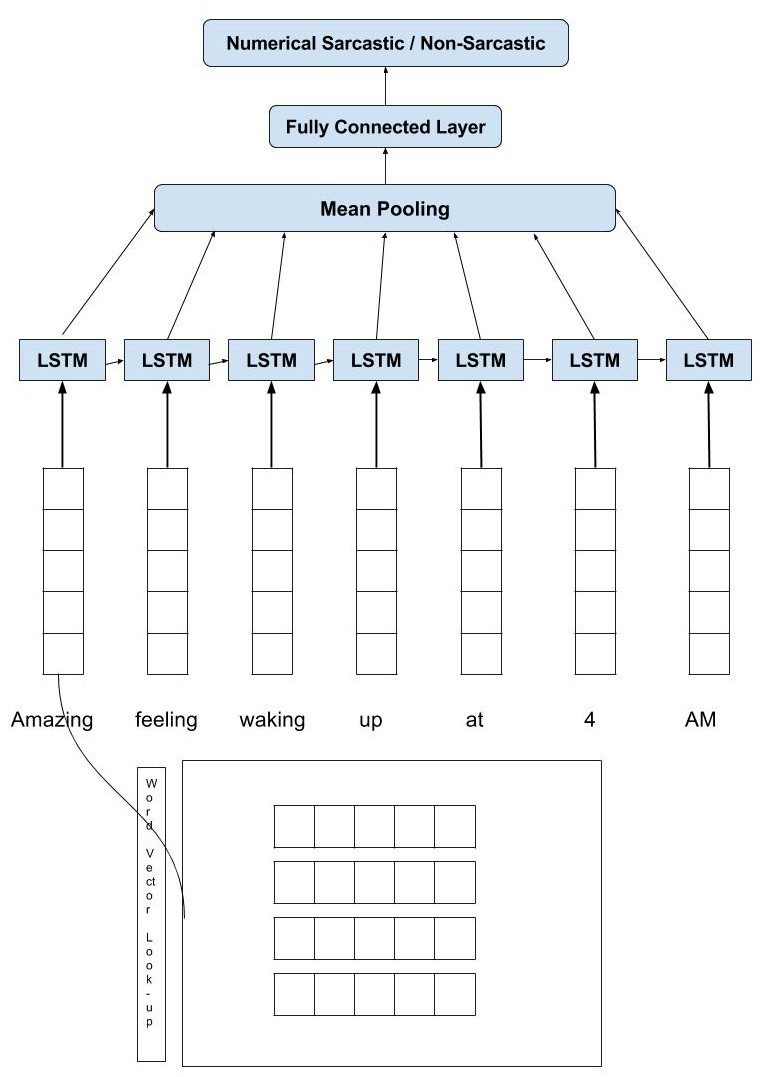}
\caption{LSTM architecture for Sarcasm}
\label{lstm_arch}
\end{figure}

This architecture is shown in \autoref{lstm_arch}. In order to convert the input tweet $T$ into its matrix representation $I$, embedding matrix $E$ is used as explained in \ref{cnn_ff}. This input matrix is given as input to LSTM cell one word at a time. The output from each time step is stored, on which mean-pooling operation is performed to get the final feature vector of the tweet. This feature vector is passed to the fully connected layer and model is trained by minimizing binary cross-entropy error.
\subsubsection{CNN-LSTM-FF Model}
\begin{figure}[!htb]
\centering
\includegraphics[width=7cm, height=6cm]{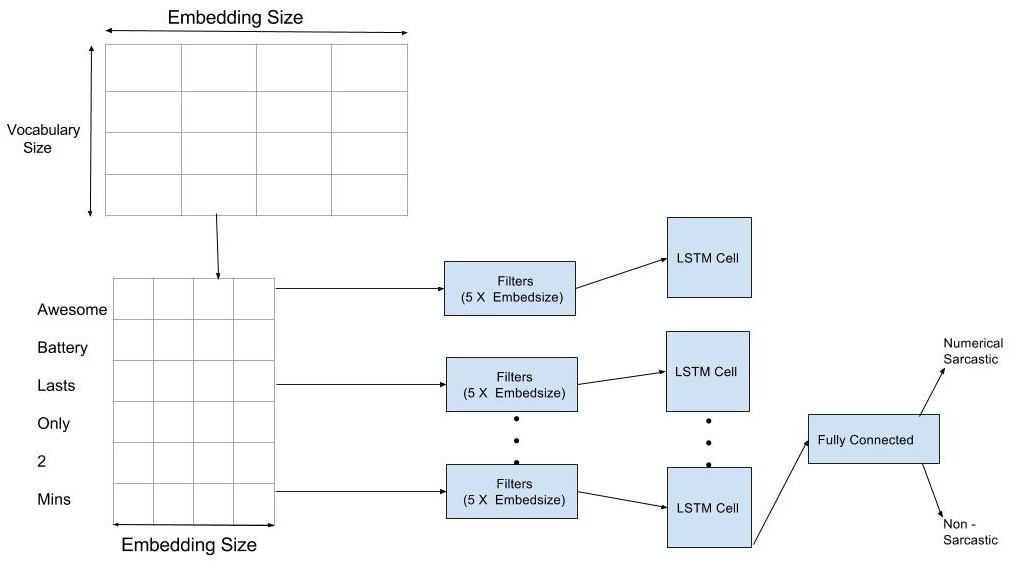}
\caption{CNN Followed by LSTM Network}
\label{fig:cnn_lstm}
\end{figure}

This architecture is shown in Figure \ref{fig:cnn_lstm}. The input matrix representation for the Tweet $T$ is obtained from the Embedding matrix $E$ as described in \ref{cnn_ff}. Filters of size $5 X d$, where d is the tweet word embedding dimension, slides over the input matrix $I$ of the tweet in order to extract the features. We passed the output of the convolutional network through a pooling layer and max-pooling is used with size 4. All the filters are of same dimension and after performing pooling operation over their outputs, we obtained a concatenated feature matrix denoted as:
\begin{center}
$C = [c_1;c_2;.......c_n]$
\end{center}
where $n$ in $c_n$ denotes the total number of filters used in the architecture. Feature matrix $C \in {\rm I\!R}^{l \times n} $, each $c_i\in {\rm I\!R}^{l}$, where l is the dimension obtained after pooling operation. Let $x_j \in  {\rm I\!R}^{1 \times n}$ is vector obtained from matrix $C$. Vector $x_j$ is the input for the LSTM cell at $j^{th}$ timestep and the LSTM cell runs for $l$ timesteps taking different input obtained from matrix $C$, at each timestep. At the end of $l^{th}$ timestep, output from the LSTM cell act as input for the fully connected layer and training is performed similar to other architectures, i.e., by minimizing binary cross-entropy loss. 

\section{Experiment Setup}
\label{section:expsetup}

We use three datasets for performing experiments which are described in Table \ref{table:dataset} and their creation details are described below: \\
(A) \textbf{Dataset-1:} To create this dataset, we extract tweets from Twitter-API (\url{https://dev.twitter.com}). The tweets containing hashtags \#sarcasm, \#sarcastic, \#BeingSarcastic are labeled as sarcastic, while those with  \#nonsarcasm, \#notsarcastic are labeled as the non-sarcastic. We remove URLs, duplicate tweets, retweets, Username and other Non-ASCII characters in these tweets.\\ 
(B) \textbf{Dataset-2:} From Dataset-1, we retain only the tweets that contain numerical characters 
to create Dataset-2. Additional processing is performed to remove irrelevant tweets, like the ones which 
contains alphabet or special character adjacent to a number like \textit{Model34d, 4s,} $<$$3$ (heart smiley) etc. 
We divide this dataset into Training set containing \textbf{8681} Non-Sarcastic tweets and \textbf{8681} Numerical Sarcastic tweets.\\
(C) \textbf{Dataset-3:} For Deep learning experiments, we need a bigger dataset, so we created Dataset-3 from dataset-1, whose details are given in Table \ref{table:dataset}. Since there were no more Numeric-Sarcastic tweets we added more Non-Sarcastic tweets to increase dataset size. \\
(D) \textbf{Test Data:} Test set details are given in Table \ref{table:dataset}. This dataset is used to evaluate previous approaches as well as all the approaches mentioned in this paper.

\begin{table}[h]
  
 \centering
    \def\arraystretch{1} 
   \resizebox{\columnwidth}{!}{%
   \begin{tabular}{|c|c|c|c|}
   \hline
   
  \multirow{1}{*}{ \textbf{Dataset-1} } 
           & 100000 \textbf{(Sarcastic)}  & 250000 \textbf{(Non-Sarcastic)}\\ \hline
   \multirow{1}{*}{ \textbf{Dataset-2} } 
			& 8681 \textbf{(Numeric Sarcastic)}  & 8681 \textbf{(Non-Sarcastic)}  \\ \hline
  \multirow{1}{*}{ \textbf{Dataset-3} } 
			&  8681 \textbf{(Numeric Sarcastic)}  & 42107 \textbf{(Non-Sarcastic)}  \\ \hline
 \multirow{1}{*}{ \textbf{Test Data} } 
			& 1843 \textbf{(Numeric Sarcastic)}  & 8317 \textbf{(Non-Sarcastic)}  \\ \hline    
\end{tabular}}
  \caption{ Description of Datasets}
  \label{table:dataset}
\end{table}
\vspace{-1em}

We re-implement work reported by \citet{buschmeier2014impact}, \citet{gonzalez2011identifying}, \citet{liebrecht2013perfect} and \citet{joshi2015harnessing} for Sarcasm detection, to show the degradation in performance for Numeric-Sarcastic Dataset. We train classifiers for the features introduced by these approaches, using SVM$^{perf}$ by \citet{joachims2006training} with RBF kernel. We compare their performance against our  approaches, and report the average 5-fold cross-validation values in the next section.
 
We train the SVM with RBF kernel and $c = 1.0$ using grid-search, Random-forest with $number$ $of$ $estimators$ = $10$ and KNN with neighbors $K=3$ using scikit\footnote{http://scikit-learn.org/stable/modules} using features as described in section \ref{ML} and using the same dataset, i.e., Dataset-2 which is also used in Rule-based approach. The test data that we use is the same as already described in \autoref{table:dataset}.

From Dataset-3, we calculated the max length of the tweet as $36$ words so we padded all the shorter tweets by special PAD character. In all deep learning experiments, we initialized embedding matrix $E$ first by random values and then by pre-trained word embeddings and tried to investigate results with different size word embeddings. 

For CNN-FF Model, we use $128$ number of filters each of size $3$, $4$ and $5$, i.e., total $128 \times 3$ filters. Drop-out probability for this implementation is $0.5$ and training is performed by applying mini-batch gradient descent. In the pooling layer, we perform max-over time pooling over the output from each filter.

For LSTM-FF model, training is performed by applying batch-gradient descent with Adagrad optimizer having learning rate as 0.3. We also investigated with different hidden unit dimension as $20,40$ and $128$ and drop-out probability of $0.25$.

For CNN-LSTM-FF Model, number of filters are 64, each with the same dimension, i.e., $5 \times Embedding size$. We investigated the results with different size embeddings and drop-out probability is $0.25$ for this architecture. All the deep learning experiments were implemented using tensorflow \shortcite{tensor}.

\section{Results}
\label{sec:res}
In this section, we evaluate our approaches to detect sarcasm in numerical portions of text. 
\begin{table}[h]
\footnotesize
  \centering
    \def\arraystretch{1} 
    \resizebox{\columnwidth}{!}{%
   \begin{tabular}{|c|c|c|c|}
   \hline
    \textbf{Approach}  & \textbf{Dataset-1} & \textbf{Dataset-2} \\ \hline
  \multirow{1}{*}{\bfseries \citet{buschmeier2014impact} } 
           & 0.69   & 0.16 \\ \hline
   \multirow{1}{*}{\bfseries \citet{gonzalez2011identifying} } 
            & 0.68    & 0.15                  \\ \hline 
    \multirow{1}{*}{\bfseries \citet{liebrecht2013perfect} } 
       & 0.67  & 0.17 \\ \hline
    \multirow{1}{*}{\bfseries \citet{joshi2015harnessing} } 
       & 0.72   & 0.25 \\ \hline
 \end{tabular}}
  \caption{Overall F1-score degradation of previous approaches on both Datasets.}
  \label{table1}
\end{table}
\vspace{-1em}

Table \ref{table1} evaluates the performance of four previous approaches on Dataset-1 and on Dataset-2. We see that three of the past four works give an F1-score that is close to each other on Dataset-2. On Dataset-1 and Dataset-2, the best F1-score of \textbf{0.72} and \textbf{0.25} respectively are obtained by using features from \citet{joshi2015harnessing}, there is a \textbf{degradation of 47\% on Dataset-2} which contains only numerical tweets. This degradation clearly shows that these past approaches are not able to capture the sarcasm that arise due to numbers in the text. All the past 4 approaches are build to detect the normal sarcasm in which the incongruity arises due to text. When the incongruity arises due to numbers these approaches gets degraded in their performance as shown in Table \ref{table1}. This clearly shows that there is a need to develop a system that is able to capture numerical sarcasm. 
\begin{table}[h!]
\footnotesize
  \centering
  \def\arraystretch{1} 
  \resizebox{\columnwidth}{!}{%
   \begin{tabular}{|>{\bfseries}c|*{3}{c|}}\hline
    \multirow{1}{*}{\bfseries Features} & \multicolumn{1}{c|}{\bfseries P } &\multicolumn{1}{c|}{\bfseries R} &\multicolumn{1}{c|}{\bfseries F1-score} \\\cline{1-4}
    
  \multirow{1}{*}{\bfseries {S + P}} 
    &  \textbf{0.79}     & \textbf{0.67}              & \textbf{0.70}         \\ \hline
  \multirow{1}{*}{\bfseries {S + P +  E + Number Value}} 
    &  0.78     & 0.64              & 0.68         \\ \hline
    \multirow{1}{*}{\bfseries {S + P + E + Number Value + Number Unit}} 
    &  0.77     & 0.62              & 0.66         \\ \hline
    \multirow{1}{*}{\bfseries {300-D Tweet Embedding}} 
    &  \textbf{0.87}     & \textbf{0.82}              & \textbf{0.83}        \\ \hline
 \end{tabular}}
  \caption{Overall F1-Score obtained through SVM Classifier}
  \label{table4}
\end{table}

\vspace{-1em}
\begin{table}[h!]
\footnotesize
  \centering
  \def\arraystretch{1} 
  \resizebox{\columnwidth}{!}{%
   \begin{tabular}{|>{\bfseries}c|*{3}{c|}}\hline
    \multirow{1}{*}{\bfseries Features} & \multicolumn{1}{c|}{\bfseries P } &\multicolumn{1}{c|}{\bfseries R} &\multicolumn{1}{c|}{\bfseries F1-score} \\\cline{1-4}
    
  \multirow{1}{*}{\bfseries {S + P}} 
    &  0.77     & 0.61              & 0.65         \\ \hline
  \multirow{1}{*}{\bfseries {S + P + E + Number Value}} 
    &  \textbf{0.79}    & \textbf{ 0.68}              & \textbf{0.71}         \\ \hline
    \multirow{1}{*}{\bfseries {S + P + E + Number Value + Number Unit}} 
    &  0.77     & 0.61              & 0.65         \\ \hline
    \multirow{1}{*}{\bfseries {50-D Tweet Embedding}} 
    &  \textbf{0.84}     & \textbf{0.70}              & \textbf{0.74}        \\ \hline
 \end{tabular}}
  \caption{Overall F1-Score obtained through KNN Classifier }
  \label{table5}
\end{table}

\vspace{-1em}
\begin{table}[h!]
\footnotesize
  \centering
  \def\arraystretch{1} 
  \resizebox{\columnwidth}{!}{%
   \begin{tabular}{|>{\bfseries}c|*{3}{c|}}\hline
    \multirow{1}{*}{\bfseries Features} & \multicolumn{1}{c|}{\bfseries P } &\multicolumn{1}{c|}{\bfseries R} &\multicolumn{1}{c|}{\bfseries F1-score} \\\cline{1-4}
    
  \multirow{1}{*}{\bfseries {S + P}} 
    &  0.78     & 0.65              & 0.69         \\ \hline
  \multirow{1}{*}{\bfseries {S + P + E + Number Value}} 
    &  0.79     & 0.68              & 0.71         \\ \hline
    \multirow{1}{*}{\bfseries {S + P + E + Number Value + Number Unit}} 
    &  \textbf{0.81}     & \textbf{0.72}              & \textbf{0.75}         \\ \hline
    \multirow{1}{*}{\bfseries {100-D Tweet Embedding}} 
    &  \textbf{0.85}     & \textbf{0.80}              & \textbf{0.82}        \\ \hline
 \end{tabular}}
  \caption{Overall F1-Score obtained through Random Forest Classifier }
  \label{table6}
\end{table}

\begin{table*}[h!]
\fontsize{4}{5.1}\selectfont
\centering
  \label{table:all}
  \def\arraystretch{1} 
  \resizebox{\textwidth}{!}{%
   \begin{tabular}{c|c|c|c|c|c|c|c|c|c}\hline
     \multirow{2}{*}{\textbf{Approaches}} & \multicolumn{3}{c|}{\bfseries Precision } 
     &\multicolumn{3}{c|}{\bfseries Recall}  &\multicolumn{3}{c}{\bfseries F-score} \\  \cline{2-10}

 & \textbf{P(1)} & \textbf{P(0)} & \textbf{P(avg)}& \textbf{R(1)} & \textbf{R(0)} & \textbf{R(avg)} & \textbf{F(1)} & \textbf{F(0)} & \textbf{F(avg)} \\ \hline
 
 \multicolumn{10}{c}{\bfseries Past Approaches} \\ \hline
 
    Buschmeier et.al.         & 0.19       & 0.98             & 0.84              & 0.99  &0.07  &0.24   & 0.32  &0.13 &0.16           \\ 
  
     Liebrecht et.al. & 0.19    & 1.00             & 0.85             & 1.00 &0.07 &0.24   & 0.32 &0.13 &0.17           \\ 
   
 Gonzalez et.al.      & 0.19       & 0.96             & 0.83              & 0.99  &0.06  &0.23   & 0.32  &0.12 &0.15       \\           
   
     Joshi et.al.     & 0.20       & 1.00             & 0.86             & 1.00 &0.13  &0.29   & 0.33  &0.23 &\textbf{0.25 }
   \\ \hline
 
   \multicolumn{10}{c}{\bfseries Rule-Based Approaches} \\ \hline
  Approach-1         & 0.53    & 0.87            & 0.81             & 0.39 &0.92 &0.83   & 0.45 &0.90 &\textbf{0.82}          \\

    Approach-2 & 0.44    & 0.85            & 0.78           & 0.28 &0.92 &0.81   & 0.34 &0.89 &0.79         \\ \hline 
   
  \multicolumn{10}{c}{\bfseries Machine-Learning Approaches} \\ \hline
  SVM        & 0.50    & 0.95  & 0.87             
  				& 0.80   &0.82  &0.82  
                & 0.61   &0.88   &\textbf{0.83}   \\ 
  
 KNN         & 0.36   & 0.94  & 0.84            
  			   & 0.81 &0.68 &0.70   
               & 0.50 &0.79 &0.74   \\
 
  Random Forest & 0.47    & 0.93  & 0.85             
  				& 0.74 &0.81  &0.80   
                & 0.57 &0.87  &0.82   \\ \hline
   
    \multicolumn{10}{c}{\bfseries Deep-Learning Approaches} \\ \hline
    
   \textbf{CNN-FF}          & \textbf{0.88}    & \textbf{0.94}            & \textbf{0.93}           & \textbf{0.71} & \textbf{0.98} & \textbf{0.93}   & \textbf{0.79} & \textbf{0.96} & \textbf{0.93}       
   \\ 
    CNN-LSTM-FF           & 0.82    & 0.94 & 0.92             & 0.72 &0.96 &0.92   & 
 0.77 &0.95 &0.92	          \\

   LSTM-FF     & 0.76    & 0.93  & 0.90  
            & 0.68 &0.95 &0.90  
             & 0.72 &0.94 &0.90           \\ \hline  
     
  \end{tabular}} 

   \caption{ Comparison Table for Classification Results of different approaches on Test Dataset. Subscripts 0 and 1 denotes Non-Sarcastic and Numeric-Sarcastic Class respectively}
  \label{table:all}
\end{table*}



\vspace{-1em}

The results of Rule-based approaches are shown Table \ref{table:all}, along with results of other approaches. We see that among the 2 rule based approach, Approach-1 performs better with an F1-score of \textbf{0.82}.

The evaluation of machine learning experiments is done using different combination of features, i.e., Sentiment  (S), Punctuation (P), Emoticon based (E), Number value and Number Unit. We have also investigated with different dimensional $(25$-D, $50$-D$...300$-D) tweet embeddings as features. We train SVM classifier and reported the results in Table \ref{table4}. Results show that SVM give best F1-score of \textbf{0.83} with \textbf{300-D} Tweet word embedding . We train KNN classifier and results are reported in \ref{table5}. This classifiers also give best result with tweet-embedding as feature but of different dimension, i.e., \textbf{50-D}. Table \ref{table6} shows the result of Random Forest classifier. We observe from the table that as we append more features like number vale and number unit to S, P and E the F1-score becomes \textbf{0.75}. Finally, when we give \textbf{100-D} tweet embedding as input, the best F1-score is \textbf{0.82}.  

As mentioned in section \ref{section:expsetup}, we performed deep learning experiments by initializing embedding matrix in two settings, the best result for all the deep-learning experiments are obtained by initializing with the pre-trained tweet word embeddings. Refer Table \ref{table:all} for deep learning experiments results. For CNN-FF model, the F1-score of \textbf{0.93} is obtained with embedding size of \textbf{250-D}. For CNN-LSTM-FF model, the F1-score of \textbf{0.91} with 200-D embedding size and for LSTM-FF model, we obtain the F1-score of \textbf{0.90} with \textbf{25-D} embedding size.

Table \ref{table:all} compares the results for all the implemented approaches and also the past approaches. The best overall F1-score of \textbf{0.93} is obtained by CNN-FF Model. We see an \textbf{improvement of 68\%} in F1-score against the best performing past approach of \citet{joshi2015harnessing}.




\section{Error Analysis}
\label{sec:erroranal}
We classify the errors by our approaches into the following categories:
\begin{itemize}
  \setlength\itemsep{0em}
\item \textbf{Unit Mismatch/Unit Missing:} There are some tweets in which the number unit is present in very informal way like min for minutes and hr for hours. So these types of unit created problem while performing the unit matching test. In some tweets the number unit is missing for example- \textit{`i love waking up at 545'}. Unit is not present as well as the time is present in wrong format.
\item \textbf{Presence of multiple numbers:} Some tweets contain multiple numbers and this makes it more challenging to identify sarcasm due to numbers. For example- \textit{` \$34.04 for a 10 mile trip that takes 19 minutes? that makes sense'}.
\item \textbf{Sarcasm due to text but not number:} Some tweets contain number but they are sarcastic not because of the presence of number. For example- \textit{`First asthma attack in 6 years. forgot how much fun they are'}, in this tweet the sarcasm is arising because of the word ``fun" present at the end of the tweet. 
\end{itemize}


\section{Conclusion \& Future Work}
\label{sec:concl}
Our paper presents the novel idea for identifying sarcasm that arises due to the presence of numerical portions in the tweets. Numerical sarcasm is a special case of sarcasm where incongruity arises between textual and numerical content. It shows the degradation in performance of the four past approaches over the dataset containing numerical sarcastic tweets. We further present Rule based, Machine learning and Deep learning approaches for numerical sarcasm detection and obtains best overall F1-score of 0.93 from CNN-FF model. In this work we try to build a system to detect numerical sarcasm in the tweets. Our work opens a new avenue in sarcasm detection as previous approaches are unable to capture numerical sarcasm because of their ability to capture the cues for normal sarcasm. In future, all the past approaches for sarcasm detection can benefit with out work with by  separating the normal sarcasm from numerical sarcasm and improve performance.


\bibliography{ijcnlp2017}

\begin{thebibliography}{}
\expandafter\ifx\csname natexlab\endcsname\relax\def\natexlab#1{#1}\fi

\bibitem[{Abadi et~al.(2016)Abadi, Barham, Chen, Chen, Davis, Dean, Devin,
  Ghemawat, Irving, Isard, Kudlur, Levenberg, Monga, Moore, Murray, Steiner,
  Tucker, Vasudevan, Warden, Wicke, Yu, and Zheng}]{tensor}
Martin Abadi, Paul Barham, Jianmin Chen, Zhifeng Chen, Andy Davis, Jeffrey
  Dean, Matthieu Devin, Sanjay Ghemawat, Geoffrey Irving, Michael Isard,
  Manjunath Kudlur, Josh Levenberg, Rajat Monga, Sherry Moore, Derek~G. Murray,
  Benoit Steiner, Paul Tucker, Vijay Vasudevan, Pete Warden, Martin Wicke, Yuan
  Yu, and Xiaoqiang Zheng. 2016.
\newblock Tensorflow: A system for large-scale machine learning.
\newblock In {\em 12th USENIX Symposium on Operating Systems Design and
  Implementation (OSDI 16)\/}. pages 265--283.

\bibitem[{Amir et~al.(2016)Amir, Wallace, Lyu, and Silva}]{amir2016modelling}
Silvio Amir, Byron~C Wallace, Hao Lyu, and Paula Carvalho M{\'a}rio~J Silva.
  2016.
\newblock Modelling context with user embeddings for sarcasm detection in
  social media.
\newblock {\em arXiv preprint arXiv:1607.00976\/} .

\bibitem[{Bouazizi and Ohtsuki(2016)}]{bouazizi2016pattern}
Mondher Bouazizi and Tomoaki~Otsuki Ohtsuki. 2016.
\newblock A pattern-based approach for sarcasm detection on twitter.
\newblock {\em IEEE Access\/} 4:5477--5488.

\bibitem[{Buschmeier et~al.(2014)Buschmeier, Cimiano, and
  Klinger}]{buschmeier2014impact}
Konstantin Buschmeier, Philipp Cimiano, and Roman Klinger. 2014.
\newblock An impact analysis of features in a classification approach to irony
  detection in product reviews.
\newblock In {\em Proceedings of the 5th Workshop on Computational Approaches
  to Subjectivity, Sentiment and Social Media Analysis\/}. pages 42--49.

\bibitem[{Carvalho et~al.(2009)Carvalho, Sarmento, Silva, and
  De~Oliveira}]{carvalho2009clues}
Paula Carvalho, Lu{\'\i}s Sarmento, M{\'a}rio~J Silva, and Eug{\'e}nio
  De~Oliveira. 2009.
\newblock Clues for detecting irony in user-generated contents: oh...!! it's so
  easy;-.
\newblock In {\em Proceedings of the 1st international CIKM workshop on
  Topic-sentiment analysis for mass opinion\/}. ACM, pages 53--56.

\bibitem[{Davidov et~al.(2010)Davidov, Tsur, and Rappoport}]{davidov2010semi}
Dmitry Davidov, Oren Tsur, and Ari Rappoport. 2010.
\newblock Semi-supervised recognition of sarcastic sentences in twitter and
  amazon.
\newblock In {\em Proceedings of the fourteenth conference on computational
  natural language learning\/}. Association for Computational Linguistics,
  pages 107--116.

\bibitem[{Ghosh and Veale(2016)}]{ghosh2016fracking}
Aniruddha Ghosh and Tony Veale. 2016.
\newblock Fracking sarcasm using neural network.
\newblock In {\em Proceedings of NAACL-HLT\/}. pages 161--169.

\bibitem[{Gibbs(1986)}]{gibbs1986psycholinguistics}
Raymond~W Gibbs. 1986.
\newblock On the psycholinguistics of sarcasm.
\newblock {\em Journal of Experimental Psychology: General\/} 115(1):3.

\bibitem[{Gonz{\'a}lez-Ib{\'a}nez et~al.(2011)Gonz{\'a}lez-Ib{\'a}nez, Muresan,
  and Wacholder}]{gonzalez2011identifying}
Roberto Gonz{\'a}lez-Ib{\'a}nez, Smaranda Muresan, and Nina Wacholder. 2011.
\newblock Identifying sarcasm in twitter: a closer look.
\newblock In {\em Proceedings of the 49th Annual Meeting of the Association for
  Computational Linguistics: Human Language Technologies: Short Papers-Volume
  2\/}. Association for Computational Linguistics, pages 581--586.

\bibitem[{Hochreiter and Schmidhuber(1997)}]{lstm_hochreiter}
Sepp Hochreiter and J{\"u}rgen Schmidhuber. 1997.
\newblock Long short-term memory.
\newblock {\em Neural computation\/} 9(8):1735--1780.

\bibitem[{Ivanko and Pexman(2003)}]{ivanko2003context}
Stacey~L Ivanko and Penny~M Pexman. 2003.
\newblock Context incongruity and irony processing.
\newblock {\em Discourse Processes\/} 35(3):241--279.

\bibitem[{Joachims(2006)}]{joachims2006training}
Thorsten Joachims. 2006.
\newblock Training linear svms in linear time.
\newblock In {\em Proceedings of the 12th ACM SIGKDD international conference
  on Knowledge discovery and data mining\/}. ACM, pages 217--226.

\bibitem[{Joshi et~al.(2016{\natexlab{a}})Joshi, Bhattacharyya, and
  Carman}]{joshi2016automatic}
Aditya Joshi, Pushpak Bhattacharyya, and Mark~James Carman. 2016{\natexlab{a}}.
\newblock Automatic sarcasm detection: A survey.
\newblock {\em arXiv preprint arXiv:1602.03426\/} .

\bibitem[{Joshi et~al.(2015)Joshi, Sharma, and
  Bhattacharyya}]{joshi2015harnessing}
Aditya Joshi, Vinita Sharma, and Pushpak Bhattacharyya. 2015.
\newblock Harnessing context incongruity for sarcasm detection.
\newblock In {\em ACL (2)\/}. pages 757--762.

\bibitem[{Joshi et~al.(2016{\natexlab{b}})Joshi, Tripathi, Bhattacharyya, and
  Carman}]{joshi2016harnessing}
Aditya Joshi, Vaibhav Tripathi, Pushpak Bhattacharyya, and Mark Carman.
  2016{\natexlab{b}}.
\newblock Harnessing sequence labeling for sarcasm detection in dialogue from
  tv series ‘friends’.
\newblock {\em CoNLL 2016\/} page 146.

\bibitem[{Liebrecht et~al.(2013)Liebrecht, Kunneman, and van~den
  Bosch}]{liebrecht2013perfect}
CC~Liebrecht, FA~Kunneman, and APJ van~den Bosch. 2013.
\newblock The perfect solution for detecting sarcasm in tweets\# not .

\bibitem[{Mikolov et~al.(2013)Mikolov, Sutskever, Chen, Corrado, and
  Dean}]{mikolov:2013}
Tomas Mikolov, Ilya Sutskever, Kai Chen, Greg Corrado, and Jeffrey Dean. 2013.
\newblock Distributed representations of words and phrases and their
  compositionality .

\bibitem[{Poria et~al.(2016)Poria, Cambria, Hazarika, and
  Vij}]{poria2016deeper}
Soujanya Poria, Erik Cambria, Devamanyu Hazarika, and Prateek Vij. 2016.
\newblock A deeper look into sarcastic tweets using deep convolutional neural
  networks.
\newblock {\em arXiv preprint arXiv:1610.08815\/} .

\bibitem[{Rajadesingan et~al.(2015)Rajadesingan, Zafarani, and
  Liu}]{rajadesingan2015sarcasm}
Ashwin Rajadesingan, Reza Zafarani, and Huan Liu. 2015.
\newblock Sarcasm detection on twitter: A behavioral modeling approach.
\newblock In {\em Proceedings of the Eighth ACM International Conference on Web
  Search and Data Mining\/}. ACM, pages 97--106.

\bibitem[{Riloff et~al.(2013)Riloff, Qadir, Surve, De~Silva, Gilbert, and
  Huang}]{riloff2013sarcasm}
Ellen Riloff, Ashequl Qadir, Prafulla Surve, Lalindra De~Silva, Nathan Gilbert,
  and Ruihong Huang. 2013.
\newblock Sarcasm as contrast between a positive sentiment and negative
  situation.
\newblock In {\em EMNLP\/}. volume~13, pages 704--714.

\bibitem[{Tepperman et~al.(2006)Tepperman, Traum, and
  Narayanan}]{tepperman2006yeah}
Joseph Tepperman, David~R Traum, and Shrikanth Narayanan. 2006.
\newblock " yeah right": sarcasm recognition for spoken dialogue systems.
\newblock In {\em INTERSPEECH\/}.

\bibitem[{Utsumi(2000)}]{utsumi2000verbal}
Akira Utsumi. 2000.
\newblock Verbal irony as implicit display of ironic environment:
  Distinguishing ironic utterances from nonirony.
\newblock {\em Journal of Pragmatics\/} 32(12):1777--1806.

\bibitem[{Zhang et~al.(2016)Zhang, Zhang, and Fu}]{zhang2016tweet}
Meishan Zhang, Yue Zhang, and Guohong Fu. 2016.
\newblock Tweet sarcasm detection using deep neural network.
\newblock In {\em Proceedings of the 26th International Conference on
  Computational Linguistics\/}. pages 2449--2460.

\end{thebibliography}
\bibliographystyle{ijcnlp2017}

\end{document}